\documentclass[conference]{IEEEtran}
\IEEEoverridecommandlockouts
\usepackage{url}
\usepackage{amsmath,amsfonts,bm}

\def\eqref#1{equation~\ref{#1}}
\def\1{\bm{1}}

\def\vh{{\bm{h}}}

\def\vm{{\bm{m}}}

\def\vs{{\bm{s}}}

\def\vv{{\bm{v}}}

\def\vz{{\bm{z}}}

\def\mH{{\bm{H}}}

\DeclareMathAlphabet{\mathsfit}{\encodingdefault}{\sfdefault}{m}{sl}
\SetMathAlphabet{\mathsfit}{bold}{\encodingdefault}{\sfdefault}{bx}{n}

\def\sA{{\mathbb{A}}}

\usepackage{amsmath,amssymb,amsfonts}
\usepackage{todonotes} %ToDo notes
\usepackage{verbatim} %comments
\usepackage{subcaption}

\usepackage{algorithmic}
\usepackage{graphicx}
\usepackage{xcolor}
\usepackage{hyperref}

\usepackage{multirow}
\usepackage{flushend}
\usepackage[noadjust]{cite}

\def\BibTeX{{\rm B\kern-.05em{\sc i\kern-.025em b}\kern-.08em
    T\kern-.1667em\lower.7ex\hbox{E}\kern-.125emX}}

\begin{document}

\title{HyperEmbed: 
Tradeoffs Between Resources and Performance in NLP Tasks with Hyperdimensional Computing Enabled Embedding of  $n$-gram Statistics\\
\thanks{$^*$ These authors have contributed equally.
\newline
The work of DK was supported by the European Union's Horizon 2020 Research and Innovation Programme under the Marie Skłodowska-Curie Individual Fellowship Grant Agreement 839179 and in part by the DARPA's VIP (Super-HD Project) and AIE (HyDDENN Project) programs.
}
}

\author{
\IEEEauthorblockN{Pedro Alonso$^*$}
\IEEEauthorblockA{\textit{EISLAB} \\
\textit{Luleå University} \\ \textit{of Technology}\\
Luleå, Sweden \\
pedro.alonso@ltu.se}
\and
\IEEEauthorblockN{Kumar Shridhar$^*$}
\IEEEauthorblockA{\textit{Department of} \\ \textit{Computer Science} \\
\textit{ETH Zürich}\\
Zürich, Switzerland \\
shridhar.kumar@inf.ethz.ch}
\and
\IEEEauthorblockN{Denis Kleyko$^*$}
\IEEEauthorblockA{\textit{UC Berkeley} \\
Berkeley, USA \\
\textit{Research Institutes of Sweden}\\
Kista, Sweden \\
denkle@berkeley.edu}
\and
\IEEEauthorblockN{Evgeny Osipov}
\IEEEauthorblockA{\textit{DCC} \\
\textit{Luleå University} \\ \textit{of Technology}\\
Luleå, Sweden \\
evgeny.osipov@ltu.se}
\and
\IEEEauthorblockN{Marcus Liwicki}
\IEEEauthorblockA{\textit{EISLAB} \\
\textit{Luleå University} \\ \textit{of Technology}\\
Luleå, Sweden \\
marcus.liwicki@ltu.se}
}

\maketitle

\begin{abstract}
Recent advances in Deep Learning have led to a significant performance increase on several NLP tasks,
however, the models become more and more computationally demanding.
Therefore, this paper tackles the domain of computationally efficient algorithms for NLP tasks. 
In particular, it investigates distributed representations of $n$-gram statistics of texts.
The representations are formed using hyperdimensional computing enabled embedding.
These representations then serve as features, which are used as input to standard classifiers.
We investigate the applicability of the embedding on one large and three small standard datasets for classification tasks using nine classifiers. 
The embedding achieved on par $F_1$ scores while decreasing the time and memory requirements by several times compared to the conventional $n$-gram statistics, e.g., for one of the classifiers on a small dataset, the memory reduction was $6.18$ times; while train and test speed-ups were $4.62$ and $3.84$ times, respectively. 
For many classifiers on the large dataset, memory reduction was ca. $100$ times and train and test speed-ups were over $100$ times.
Importantly, the usage of distributed representations formed via hyperdimensional computing allows dissecting strict dependency between the dimensionality of the representation and $n$-gram size, thus, opening a room for tradeoffs.
\end{abstract}


\section{Introduction}
Recent work~\cite{EnergyNLP} has brought significant attention by demonstrating potential cost and environmental impact of developing and training state-of-the-art models for Natural Language Processing (NLP) tasks. 
The work suggested several countermeasures for changing the situation. 
One of them \cite{EnergyNLP} recommends a concerted effort by industry and academia to promote research of more computationally efficient algorithms.
The main focus of this paper falls precisely in this domain. 

In particular, we consider NLP systems using a well-known technique called $n$-gram statistics. 
The key idea is that hyperdimensional computing~\cite{Kanerva09} allows forming distributed representations of the conventional $n$-gram statistics~\cite{RIJHK2015}.
The use of these distributed representations, in turn, allows trading-off the performance of an NLP system (e.g., $F_1$ score) and its computational resources (i.e., time and memory). 
The main contribution of this paper is the systematic study of these tradeoffs on nine machine learning algorithms using several benchmark classification datasets.
%{\color{red}
This is the first study where the computational tradeoffs of the distributed representations of $n$-gram statistics is studied in an extensive manner on numerous datasets. 
%}
%While many works have been done lately for learning deep word representations~\cite{mikolov2013efficient}, \cite{pennington2014glove}, \cite{peters2018deep}, 
We demonstrate the usefulness of hyperdimensional computing-based embedding, which is highly time and memory efficient.
Our experiments on a well-known dataset \cite{braun-EtAl:2017:SIGDIAL} for intent classification show that it is possible to reduce memory usage by $\sim10$x and speed-up training by $\sim5$x without compromising $F_1$ score. 

%\begin{comment}
Several important use-cases are motivating the efforts towards trading-off the performance of a system against  computational resources required to achieve that performance:
high-throughput systems with an extremely large number of requests/transactions (the power of one per cent);
resource-constrained systems where computational resources and energy are scarce (edge computing);
green computing systems taking into account environmental sustainability when considering algorithms efficiency~\cite{HLAIEthics}.
%\end{comment}

%\begin{itemize}
%  \item High-throughput systems with an extremely large number of requests/transactions (the power of one per cent);
%  \item Resource-constrained systems where computational resources and energy are scarce (edge computing);
%  \item Green computing systems taking into account the aspects of environmental sustainability when considering the efficiency of algorithms.
%\end{itemize}

The paper is structured as follows. 
Section~\ref{sec:related} covers the related work.
Section~\ref{sec:outline} outlines the evaluation and describes the datasets.
The methods being used are presented in Section~\ref{sec:methods}. 
Section~\ref{sec:perf} evaluates of the experimental results. 
Discussion and conclusion are presented in Section~\ref{sec:disc}.
%Section~\ref{sec:conclusions} concludes the paper.

\section{Related Work}
\label{sec:related}
Commonly, data for NLP tasks are represented in the form of vectors, which are then used as an input to machine learning algorithms.
These representations range from dense learnable vectors to extremely sparse non-learnable vectors.
%, e.g., one-hot encodings. 
%Vector representations have been used for NLP tasks for quite some time.
Well-known examples of such representations include one-hot encodings, count-based vectors, and Term Frequency Inverse Document Frequency (TF-IDF) among others. 
Despite being very useful, non-learnable representations have their disadvantages such as resource inefficiency due to their sparsity and absence of contextual information (except for TF-IDF).
Learnable vector representations such as word embeddings (e.g., Word2Vec~\cite{mikolov2013efficient} or  GloVe~\cite{pennington2014glove}) partially address these issues by obtaining dense vectors in an unsupervised learning fashion. 
These representations are based on the distributional hypothesis:  words located nearby in a vector space should have similar contextual meaning. 
The idea has been further improved in~\cite{2016arXiv160701759J} by representing words with character $n$-grams. 
Another efficient way of representing a word is the concept of Byte Pair Encoding, which has been introduced in~\cite{gage1994new}.
The disadvantage of the learnable representations, however, is that they require pretraining involving large train corpus as well as have a large memory footprint (in order of GB). 
As an alternative to word/character embedding,~\cite{SemHash19} introduced the idea of Subword Semantic Hashing that uses a hashing method to represent subword tokens, thus, reducing the memory footprint (in order of MB) and removing the necessity of pretraining over a large corpus.
The approach has demonstrated state-of-the-art results on $3$ intent classification datasets.

The Subword Semantic Hashing, however,  relies on $n$-gram statistics for extracting the representation vector used as an input to classification algorithms. 
It is worth noting that the conventional $n$-gram statistics uses a positional representation where each position in the vector can be attributed to a particular $n$-gram.
The disadvantage of the conventional $n$-gram statistics is that the size of the vector grows exponentially with $n$.
Nevertheless, it is possible to untie the size of representation from $n$ by using distributed representations~\cite{Hinton1986}, where the information is distributed across the vector’s positions.
In particular,~\cite{RIJHK2015} suggest how to embed conventional $n$-gram statistics into a high-dimensional vector (HD vector) using the principles of hyperdimensional computing.
Hyperdimensional computing also known as Vector Symbolic Architectures~\cite{PlateBook, RachkovskijStructures2001, Kanerva09, BuildBrain, FradySDR2020} is a family of bio-inspired methods of manipulating and representing information. 
The method of embedding $n$-gram statistics into the distributed representation in the form of an HD vector has demonstrated promising results on the task of language identification while being hardware-friendly~\cite{RahimiLPHD}. 
In~\cite{Rasti2016} it was further applied to the classification of news articles and for traffic clustering in~\cite{BandaragodaTrajectoryTraffic2019}.
The method has also shown promising results~\cite{PSI19} when using HD vectors for training Self-Organizing Maps~\cite{SOMBook, KleykointSOM2019}. 
There are also other domains, which have not required the use of $n$-gram statistics, but benefited from  the use of hyperdimensional computing: biomedical signal processing~\cite{ACCESS_HRV, HDGestureIEEE}, seizure onset detection and localization~\cite{BurrelloLBPSeizure2020}, gesture recognition~\cite{TNNLS18,HD_ICRC16}, fault isolation~\cite{ACCESS_BIOFAULT}, physical activity recognition~\cite{Rasanen14}, and communications~\cite{JakimovskiCollective2012, KleykoMACOM2012, KimHDM2018}.
However, there are no previous studies comprehensively exploring tradeoffs achievable with the method on  NLP datasets with the supervised classifiers. 

\section{Evaluation outline}
\label{sec:outline}

\subsection{Classifiers and performance metrics}
\label{sec:classifiers}
To obtain the results applicable to a broad range of existing machine learning algorithms, we have performed experiments with several conventional classifiers.\footnote{
In the experiments below, we have considered the standard centralized setup where all the data was available at once to train a single model but HD vectors can be also useful in a distributed classification setup~\cite{RosatoHDDistributed2021}.
}
In particular, 
%following the experimental protocol in~\cite{SemHash19}, 
the following classifiers were studied: Ridge Classifier (Ridge), k-Nearest Neighbors (KNN), Multilayer Perceptron (MLP), Passive Aggressive (PA), Random Forest (RF), Linear Support Vector Classifier (LSVC), Stochastic Gradient Descent (SGD), Nearest Centroid (NC), and Bernoulli  Naive Bayes (BNB).\footnote{
It is worth noting that the choice of classifiers does not have to be limited to the ones used in this study. 
For example, recently Learning Vector Quantization classifier was demonstrated to learn well on HD vectors~\cite{DiaoGLVQHD2021}.
}
All the classifiers are available in the scikit-learn library~\cite{Scikit}, which was used in the experiments.

Since the main focus of this paper is the tradeoff between classification performance and computational resources, we have to define metrics for both aspects. 
The quality of the classification performance of a model will be measured by a simple and well-known metric -- $F_1$ score (please see~\cite{ROC}). 
The computational resources will be characterized by three metrics: the time it takes to train a model, the time it takes to test the trained model, and the memory, where the memory is defined as the sum of the size of input feature vectors for train and test splits as well as the size of the trained model. 
To avoid the dependencies such as particular specifications of a computer and dataset size, the train/test times and memory are reported as relative values (i.e., train/test speed-up and memory reduction), where the reference is the value obtained for the case of the conventional $n$-gram statistics.\footnote{
Note that the speed-ups reported in Section~\ref{sec:perf} do not include the time it takes to obtain the corresponding HD vectors.
Please see the discussion of this issue in Section~\ref{sec:disc}. }

\subsection{Datasets}
\label{sec:data}
Four different datasets were used to obtain the empirical results reported in this paper: the \textit{Chatbot Corpus} (Chatbot), the \textit{Ask Ubuntu Corpus} (AskUbuntu), the \textit{Web Applications Corpus} (WebApplication), and the \textit{20 News Groups Corpus} (20NewsGroups).
The first three are referred to as small datasets.
The Chatbot dataset comprises questions posed to a Telegram chatbot. The chatbot, in turn, replied the questions of the public transport of Munich.
The AskUbuntu and WebApplication datasets are questions and answers from the StackExchange. 
The 20NewsGroups dataset comprises news posts labelled into several categories.
%Each file in the dataset is one full post from a category including additional metadata such as organization, title, from, to, etc.
All datasets have predetermined train and test splits. The first three datasets~\cite{braun-EtAl:2017:SIGDIAL} are available on GitHub.\footnote{Under the Creative Commons CC BY-SA 3.0 license: \url{https://github.com/sebischair/NLU-Evaluation-Corpora}}

%\subsubsection{The Chatbot Corpus}
The Chatbot dataset consists of two intents: the \textit{(Departure Time and Find Connection)} with 206 questions. 
The corpus has a total of five different entity types \textit{(StationStart, StationDest, Criterion, Vehicle, Line)}, which were not used in our benchmarks, as the results were only for intent classification.
The samples come in English. 
Despite this, the train station names are in German, which is evident from the text where the German letters appear ({\"a},{\"o},{\"u},\ss). 
The dataset has the following data sample distribution (train/test): Departure Time ($43/35$); Find Connection ($57/71$).
%Table~\ref{tab:data distribution Chatbot} presents the data sample distribution for the Chatbot dataset.

%The dataset has the following data sample distribution (train/test): Departure Time ($43/35$); Find Connection ($57/71$).

%Table \ref{tab:data distribution AskUbuntu} shows the data distribution of AskUbuntu Corpus.

%\subsubsection{The AskUbuntu Corpus}
%The AskUbuntu dataset %consists of
%comprises five intents with the following data sample distribution (train/test): 
%Make Update ($10/37$); 
%Setup Printer ($10/13$); 
%Shutdown Computer ($13/14$); 
%Software Recommendation ($17/40$); 
%None ($3/5$).
%Thus, it includes $162$ samples in total. 

\begin{comment}
The AskUbuntu dataset comprises five intents: 
Make Update; 
Setup Printer; 
Shutdown Computer; 
Software Recommendation; 
None.
It includes $162$ samples in total. 
Please refer to Table~\ref{tab:data distribution AskUbuntu} for its data sample distribution. 
\end{comment}

The AskUbuntu dataset %consists of
comprises five intents with the following data sample distribution (train/test): 
Make Update ($10/37$); 
Setup Printer ($10/13$); 
Shutdown Computer ($13/14$); 
Software Recommendation ($17/40$); 
None ($3/5$).
Thus, it includes $162$ samples in total. 
The samples were gathered directly from the AskUbuntu platform.
Only questions with the highest scores and upvotes were considered.
%For the task of mapping the correct intent to the question, the Amazon Mechanical Turk was employed.
%Beyond the questions labelled with their intent, this dataset contains also some extra information such as author, page URL with the question, entities, answer, and the answer's author.
%It is worth noting that none of these data were used in the experiments.

%\subsubsection{The Web Applications Corpus}
%The WebApplication dataset consist of the same features and was preprocessed in a similar manner as the AskUbuntu one.

The WebApplication dataset comprises $89$ text samples of eight intents with the following distribution (train/test):
Change Password ($2/6$);  
Delete Account ($7/10$);  
Download Video ($1/0$); 
Export Data ($2/3$); 
Filter Spam ($6/14$); 
Find Alternative ($7/16$); 
Sync Accounts ($3/6$);  
None ($2/4$).

The 20NewsGroups dataset has been originally collected by Ken Lang. 
It comprises of $20$ categories.
Each category has exactly $18,846$ text samples.
Moreover, the samples of each category are split neatly into the train ($11,314$ samples) and test ($7,532$ samples) sets.
The dataset comes prepackaged with Python \textit{scikitlearn} library.

\section{Methods}
\label{sec:methods}

\subsection{Conventional \textit{n}-gram statistics}
\label{sec:conv:ngrams}

An empty vector $\displaystyle \vs$ stores $n$-gram statistics for an input text $\mathcal{D}$.
$\mathcal{D}$ consists of symbols from the alphabet of size $a$;
 $i$th position in $\displaystyle \vs$  keeps the counter of the corresponding $n$-gram $ \displaystyle \bm{\sA}_i=\langle
\mathcal{S}_1, \mathcal{S}_2, \dots, \mathcal{S}_n,  \rangle $ from the set $\displaystyle \sA$ of all unique $n$-grams; $\mathcal{S}_j$ corresponds to a symbol in $j$th position of $\displaystyle \sA_i$.
The dimensionality of $\displaystyle \vs$  equals the total number of $n$-grams in
$\displaystyle \sA$ and calculated as $a^n$.
Usually, $\displaystyle \vs$ is obtained via a single pass-through
$\mathcal{D}$ using the overlapping sliding window of size $n$. 
The value of a position in $\displaystyle \vs$ (i.e., counter) corresponding to an $n$-gram observed in the current window is incremented, i.e., $\displaystyle \vs$ summarizes how many times each  $n$-gram in $\displaystyle \sA$ was observed in $\mathcal{D}$.

\subsection{Embeddings with Subword Information}

Work by~\cite{bojanowski2017enriching} demonstrated that words' representations can be formed via learning character $n$-grams, which are then summed up to represent words. 
This method (FastText) has an advantage over the conventional word embeddings since unseen words could be better approximated as it is highly likely that some of their $n$-gram subwords have already appeared in other words. 
Therefore, each word $w$ is represented as a bag of its character $n$-gram.
Special boundary symbols ``\texttt{<}'' and ``\texttt{>}'' are added at the beginning and the end of each word. 
The word $w$ itself is added to the set of its $n$-grams, to learn a representation for each word along with character $n$-grams. 
Taking the word \emph{have} and $n=3$ as an example, $\mathcal(have)=[<ha, hav, ave, ve>, have]$. 
Formally, for a given word $w$, $\mathcal{N}_w \subset \{1, \dots, G \}$ denotes the set of $G$ $n$-grams appearing in $w$.
Each $n$-gram $g$ has an associated vector representation $\displaystyle \vz_g$.
Word $w$ is represented as the sum of the vector representations of its $n$-grams.
A scoring function $g$ is defined for each word that is represented as a set of respective $n$-grams and the context word (denoted as $c$), as:
\begin{equation*}
g(w, c) = \sum_{g \in \mathcal{N}_w} \displaystyle \vz_g^\top \displaystyle \vv_c,
\end{equation*}
where $\displaystyle \vv_c$ is the vector representation of the context word $c$.
Practically, a word is represented by its index in the dictionary and a set of its $n$-grams.

\subsection{Byte Pair Encoding}
The idea of Byte Pair Encoding (BPE) was introduced in~\cite{gage1994new}.
BPE iteratively replaces the most frequent pair of bytes in a sequence with a single, unused byte. 
It can be similarly used to merge characters or character sequences for words representations. 
A symbol vocabulary is initialized with a character vocabulary with every word represented in the form of characters, where ``$.$'' is used as the end of word symbol. 
All symbol pairs are counted iteratively and then replaced with a new symbol. 
Each operation results in a new symbol, which represents an $n$-gram.
Similarly, frequently occurring $n$-grams are eventually merged into a single symbol. 
This makes the final vocabulary size equal to the sum of initial vocabulary and number of merge operations.

\subsection{SubWord Semantic Hashing}
Subword Semantic Hashing (SemHash) is described in details in~\cite{SemHash19, SemHash13}.
%, which has been inspired by the Deep Semantic Similarity Model proposed by~\cite{shen2014learning}. 
%According to the authors, 
SemHash represents the input sentence in the form of subword tokens using a hashing method reducing the collision rate. 
These subword tokens act as features to the model and can be used as an alternative to word/$n$-gram embeddings.  
For a given input sample text $T$, e.g., \textit{``I have a flying disk''}, we split it into a list of words $t_i$.
The output of the split would look as follows: \textit{[``I'', ``have'', ``a'', ``flying'', ``disk'']}.
Each word is then passed into a prehashing function $\mathcal{H}(t_i)$.
$\mathcal{H}(t_i)$ first adds a $\#$ at the beginning and at the end of $t_i$.
Then it generates subwords via extracting $n$-grams ($n$=3) from  $\# t_i \#$, e.g., $\mathcal{H}(have)=[\#ha, hav, ave, ve\#]$.
These tri-grams are the subwords denoted as $t^j_i$, where $j$ is the index of a subword. 
$\mathcal{H}(t_i)$ is then applied to the entire text corpus to generate subwords via $n$-gram statistics. 
These subwords are used to extract features for a given text. 
% In other words, the VSM acts as a hashing function for an input text sequence. 

% \todo[inline, color=cyan]{
% DK: "These subwords are then used to create a Vector Space Model (VSM). 
% This VSM should be used to extract features for a given input text. 
% In other words, this VSM acts as a hashing function for an input text sequence." \\
% We have to think about this passage. In a way, we have not really introduced what is VSM. This might confuse a reader a lot. Maybe for the purposes of this paper it would be easier to simply say that after $\mathcal{H}(t_i)$  being applied to the entire text corpus we use $n$-gram statistics as a way to extract feature vector. 
% }

\subsection{Embedding \textit{n}-grams into an HD vector}
\label{sec:HD:embed}

Alphabet's symbols are the most basic elements of a system.
We assign each symbol `\textit{a}' with a random $d$-dimensional bipolar HD vector. 
These vectors are stored in a matrix (denoted as $\displaystyle \mH$, where $\displaystyle \mH \in [d \times a]$), which is referred to as the item memory, 
For a given symbol $\mathcal{S}$  its HD vector is denoted as $\displaystyle \mH_{\mathcal{S}} \in \{-1, +1\}^{[d \times 1]}$. 
%It is worth noting that an important property of high-dimensional spaces is that with an extremely high probability all random HD vectors are dissimilar to each other (quasi orthogonal).
To manipulate HD vectors, hyperdimensional computing defines three key operations\footnote{Please see~\cite{Kanerva09} for proper definitions and properties of hyperdimensional computing operations.} on them: bundling (denoted with $+$ and implemented via position-wise addition), binding (denoted with $\odot$ and implemented via position-wise multiplication), and permutation\footnote{$\displaystyle \rho$ is used to bind HD vector with its position.} (denoted with $\displaystyle \rho$). 
The bundling operation allows storing information in HD vectors~\cite{kleyko2016holographic};
if several copies of any HD vector are included (e.g., $2\displaystyle \mH_{\mathcal{S}_1} + \displaystyle \mH_{\mathcal{S}_2}$), the resultant HD vector is more similar to the dominating HD vector than to other components.
%{\color{red}
Since the main focus of this paper is on empirical demonstration of the usefulness of embedding n-gram statistics to HD vectors it does not go into deep analytical details of why HD vectors allow embedding the conventional $n$-gram statistics, the diligent readers are referred to~\cite{Frady17, KleykoPerceptron2020} for the relevant analysis.

It is worth mentioning, however, that intuitively the whole approach works because the embedding is done in such a way that in the projected space, two similar $n$-gram statistics (in the original space)  still remain very similar. 
%}

Three operations above allow embedding various data structures into distributed representation (HD vector): sequences~\cite{Kanerva09}, sets~\cite{KleykoABF2020}, state automata~\cite{YerxaUCBHD_FSA2018, OsipovHD_FSA2017}.
See~\cite{KleykoComputingParadigm2021} for more examples.
The embedding for $n$-gram statistics was proposed in~\cite{RIJHK2015}. 
First,  $\displaystyle \mH$ is generated for the alphabet.
A position of symbol $\mathcal{S}_j$ in $\displaystyle \sA_i$ is represented  by applying $\displaystyle \rho$ to the corresponding HD vector  $\displaystyle \mH_{\mathcal{S}_j}$ $j$ times, which is denoted as $\displaystyle \rho^{j}(\displaystyle \mH_{\mathcal{S}_j})$. 
Next, a single HD vector for $\displaystyle \sA_i$ (denoted as $\displaystyle \vm_{\displaystyle \sA_i}$) is formed via the consecutive binding of permuted HD vectors $\displaystyle \rho^{j}(\displaystyle \mH_{\mathcal{S}_j})$ representing symbols in each position $j$ of $\displaystyle \sA_i$. 

To concrete these ideas with an example consider the tri-gram  `cba', it will be mapped to its HD vector as follows: $\displaystyle  \rho^{1}(\displaystyle \mH_{\text{c}}) \odot \displaystyle \rho^{2}(\displaystyle \mH_{\text{b}}) \odot \displaystyle  \rho^{3}(\displaystyle \mH_{\text{a}}) $.  
In general, the process of forming HD vector of an $n$ can be formalized as follows: 
\noindent
\begin{equation*}
%\label{eq:hdstat} 
\displaystyle \vm_{\displaystyle \sA_i} = \prod_{j=1}^{n} \displaystyle \rho^{j}(\displaystyle \mH_{\mathcal{S}_j}), 
\end{equation*}
\noindent
where $\prod$ denotes the binding operation when applied to $n$ HD vectors.
Once it is known how to get  $\displaystyle \vm_{\displaystyle \sA_i}$, embedding the conventional $n$-gram statistics stored in $\displaystyle \vs$ (see section \ref{sec:conv:ngrams}) is straightforward. 
HD vector $\displaystyle \vh$ corresponding to $\displaystyle \vs$  is created  by bundling all $n$-grams observed in the data:
\noindent
\begin{equation*}
%\label{eq:hdstat} 
\displaystyle \vh=\sum_{i=1}^{a^n} \displaystyle \vs_i  \displaystyle \vm_{\displaystyle \sA_i} = \sum_{i=1}^{a^n} \displaystyle \vs_i   \prod_{j=1}^{n} \displaystyle \rho^{j}(\displaystyle \mH_{\mathcal{S}_j}), 
\end{equation*}
\noindent
where $\sum$ denotes the bundling operation when applied to several HD vectors. 

Let us consider an example of forming $\displaystyle \vh$ for tri-gram statistics extracted from the word ``hello''.
The tri-gram statistics includes three tri-grams: `hel', `ell', and `llo'.
So the HD vector representing the statistics is formed as: \noindent
\begin{equation*}
%\label{eq:hdstat} 
\begin{split}
\displaystyle \vh & =  
\displaystyle  \rho^{1}(\displaystyle \mH_{\text{h}}) \odot \displaystyle \rho^{2}(\displaystyle \mH_{\text{e}}) \odot \displaystyle  \rho^{3}(\displaystyle \mH_{\text{l}})
+ \\
&+\displaystyle  \rho^{1}(\displaystyle \mH_{\text{e}}) \odot \displaystyle \rho^{2}(\displaystyle \mH_{\text{l}}) \odot \displaystyle  \rho^{3}(\displaystyle \mH_{\text{l}})
+ \\
&+\displaystyle  \rho^{1}(\displaystyle \mH_{\text{l}}) \odot \displaystyle \rho^{2}(\displaystyle \mH_{\text{l}}) \odot \displaystyle  \rho^{3}(\displaystyle \mH_{\text{0}})
. 
\end{split}
\end{equation*}
\noindent
Note that $\displaystyle \vh$ is not bipolar due to the usage of the bundling operation.
In fact, the components in  $\displaystyle \vh$ will be integers in the range $[-\displaystyle\sum_{i=1}^{i=a^n}\displaystyle \vs_i,\displaystyle\sum_{i=1}^{i=a^n}\displaystyle \vs_i]$
but these extreme values are highly unlikely since  HD vectors for different $n$-grams are quasi orthogonal, which means that in the simplest (but not practical) case when all $n$-grams have the same probability the expected value of a component in $\displaystyle \vh$ is $0$. 
Also, the use of $\sum$ means that two HD vectors mapping two different $n$-gram statistics might have very different magnitudes if the number of observations in these statistics are very different, therefore, it is convenient to use the cosine  similarity between HD vectors as it neglects the magnitude.
Since there is no simple way to set a particular metric for a given machine learning algorithm (usually the Euclidean distance is used), in the experiments below we have imposed the use of the cosine  similarity implicitly by normalizing each  $\displaystyle \vh$ by its $\ell _2$ norm, thus, all $\displaystyle \vh$ had the same norm and their dot product was equivalent to their cosine  similarity.
It is worth noting that the normalization of an HD vector could also be used as a way to keep its values in the limited range, which might be useful for efficient implementation of the conventional machine learning algorithms~\cite{KleykointSOM2019, intESN2020, intRVFL2020}, but we do not study this effect in this paper.

%}

\subsection{Motivation for the chosen baselines}

Since the primary claim in this paper is that with HD vectors, it is possible to approximate (even accurately) the results obtained with the conventional $n$-gram statistics,  the most proper baseline for classification performance comparison is the conventional $n$-gram statistics itself\footnote{
Though, we do not make any definite statements such as that the $n$-gram statistics is a superior technique for solving all NLP problems. 
The only claim is that it is a well-known technique, which is still useful for numerous problems.
}.
It is also worth mentioning that there are methods (see, e.g.,~\cite{MassiveGrams}) for making efficient data structures for storing $n$-gram statistics.
However, such approaches rely on the fact that there are clear regularities when words are used as the basic elements for $n$-gram statistic. This is not the case for character $n$-grams are used here.

In addition to the methods presented above, while designing the evaluation experiments it was considered whether word embeddings such as Word2vec~\cite{mikolov2013efficient} or GloVe~\cite{pennington2014glove}) should be used as baselines. 
It was concluded that from a computational point of view, it would be unfair neglecting the computational resources spent while training these embeddings.  
On top of this, the require quite some memory even to keep the learned embedding for each word in the dictionary. 
Therefore, trainable word embeddings are not part of the baseline as the resources needed to train them are significantly higher. 
One exception, however, was made for the case of FastText, which are the trainable subword embeddings. 
Please see the discussion on this matter at the end of Section~\ref{sect:results}. 

When it comes to other well-known methods such as bag of words and TF-IDF, it was decided that since the dimensionality of the input feature equals the number of words in the dictionary, the computational efficiency of both approaches would not be much better than that of the conventional $n$-gram statistics.
This assumption is correct, at least for the small datasets, where the number of unique $n$-grams is in the order of several thousand. 
At the same time, it was relevant to observe whether HD vectors could be used to embed bag of words and TF-IDF features, therefore, the experiments on the small datasets were also performed with these methods.

\subsection{Setup}
All datasets were preprocessed using the spacy library. It was used to remove stop words from data. We used the spacy model called ``en\_core\_web\_lg'' to parse the datasets. 
Also, all text samples were preprocessed by removing control characters; 
in particular, the ones in the set \textit{[Cc]}, which includes Unicode characters from \textit{U+0000} to \textit{U+009F}.
It is also worth noting that the realization of the conventional $n$-gram statistics used in the experiments was forming a model storing only $n$-grams present in the train split. 

Since the 20NewsGroups dataset is already large, it does not seem to be necessary to apply the SemHash to it, therefore, it was omitted in the experiments (i.e., SH in Table~\ref{tab:classifiers:20NewsGroups} refers to pure $n$-grams).
Last, the small datasets were augmented, making all smaller classes having the same number of samples as the largest class in the train split for that dataset. 
Using WordNet as a dictionary, nouns and verbs were swapped with their synonyms creating new sentences until all the classes for that set have the same number of samples. 
%The final distributions were already shown in Tables~\ref{tab:data distribution Chatbot}--\ref{tab:data distribution WebApplication}.

For BPE, vocabulary size of $1000$ was used for WebApplication and AskUbuntu dataset whereas a vocabulary size of $250$ was used for Chatbot dataset due to its smaller size. $n$-gram range of ($2-4$) was used with analyzer as \textit{char}. Cross-validation was set to $5$. 

For FastText, autotune validation was used to find optimal hyperparameters for all datasets. No quantization of the model was performed to prevent the compromise on model accuracy. 

When it comes to hyperparameters, in order to find optimal hyperparameters, a grid-based search was applied to three small datasets for the following classifiers: MLP, RF, and KNN.
The configuration performing best among all small datasets was chosen to be used report the results in this paper. 
Moreover, the same configuration was used for the 20NewsGroups dataset.

In the case of MLP, four different configurations of hidden layers were considered:  [(100, 50), (300, 100),(300, 200, 100), and (300, 100, 50)]; (300, 100, 50) configuration has been chosen. The maximal number of MLP iterations was set to 500.
In the case of RF, two hyperparameters were optimized number of estimators ([50, 60, 70]) and minimum samples leaf ([1, 11]); we used  50 estimators and 1 leaf.
In the case of KNN, the number of neighbors between 3 and 7 was considered; 3 neighbors were used in the experiments.  
For all the other classifiers default hyperparameter settings provided by Sklearn library were used.

The range of $n$ in the experiments with small datasets was $[2-4]$ while for the 20NewsGroups dataset it was $[2-3]$ since the number of possible $4$-grams was overwhelming. 
All results reported for small datasets were obtained by averaging across $50$ independent simulations.
In the case of the 20NewsGroups dataset, the number of simulations was decreased to $10$ due to high computational costs.
To have a fair comparison of computational resources, all results for small datasets were obtained on a dedicated laptop without involving GPUs while the results for the 20NewsGroups were obtained with a computing cluster (CPU only) without the intervention of other users.

%\section{Experimental Settings}

 \begin{figure*}[h]%[t]%[ht]
\centering
\begin{subfigure}[h]{0.98\linewidth}
\centering
   \includegraphics[width=0.99\linewidth]{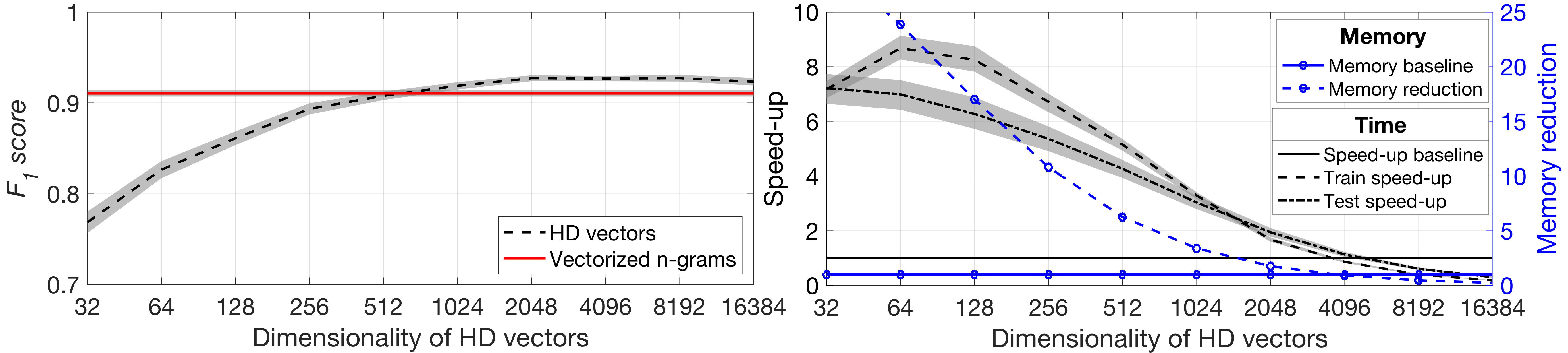}
   \caption{AskUbuntu}
   \label{fig:AskUbuntu} 
\end{subfigure}

\begin{subfigure}[h]{0.98\linewidth}
\centering
   \includegraphics[width=0.99\linewidth]{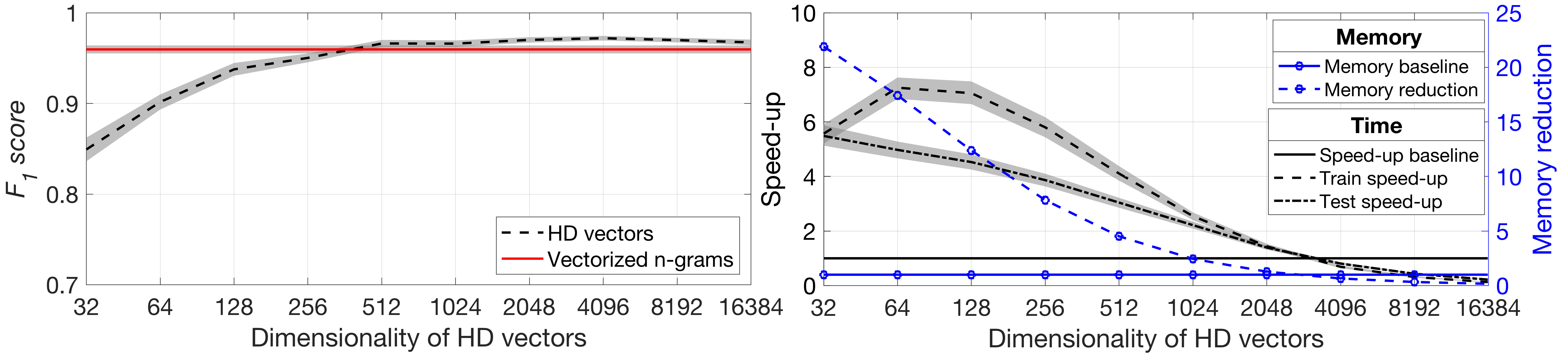}
   \caption{Chatbot}
   \label{fig:Chatbot} 
\end{subfigure}

\begin{subfigure}[h]{0.98\linewidth}
\centering
   \includegraphics[width=0.99\linewidth]{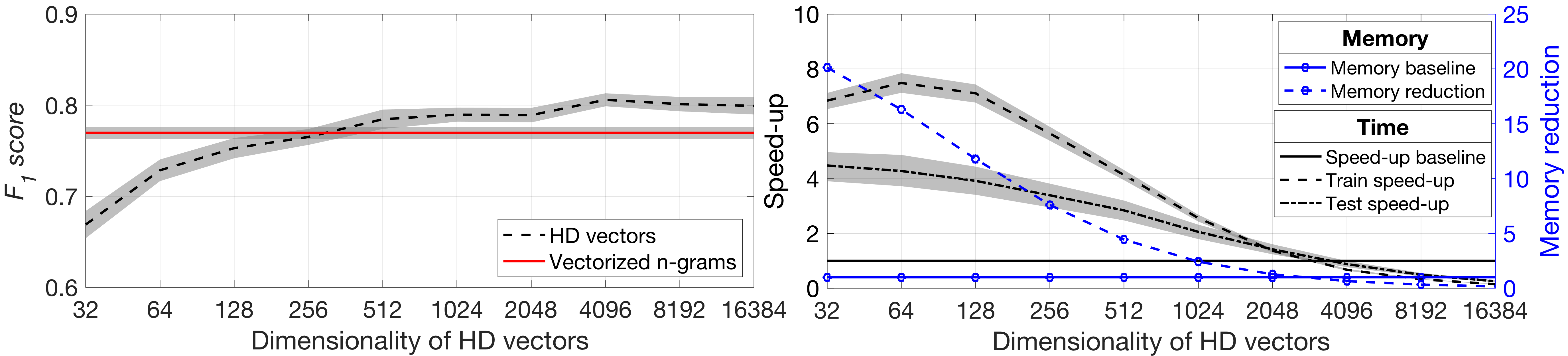}
   \caption{WebApplication}
   \label{fig:WebApplication} 
\end{subfigure}

\begin{subfigure}[h]{0.98\linewidth}
\centering
   \includegraphics[width=0.99\linewidth]{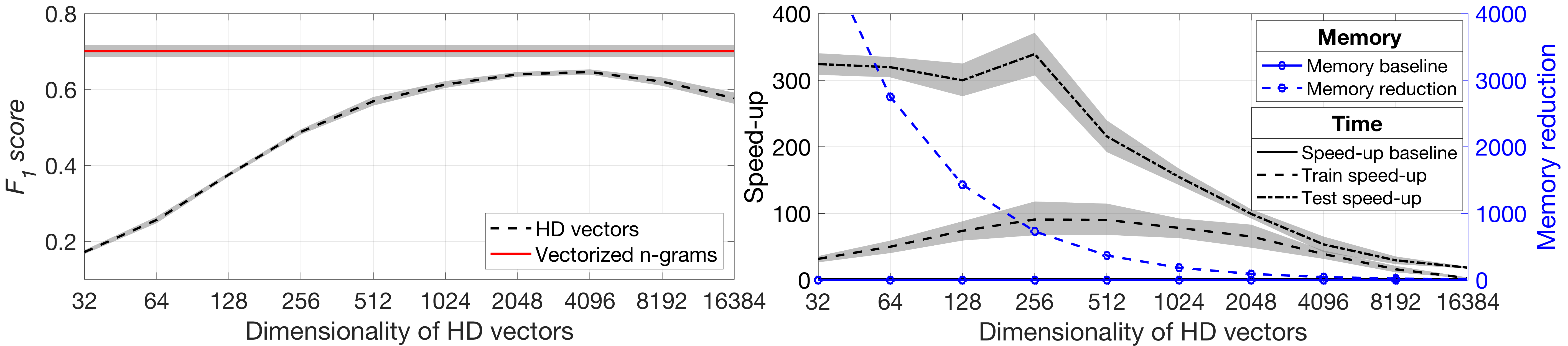}
   \caption{20NewsGroups}
   \label{fig:20NewsGroups} 
\end{subfigure}

\caption{
MLP results vs. the dimensionality of HD vectors.
}
\end{figure*}

\section{Empirical evaluation}
\label{sec:perf}

\subsection{Results}
\label{sect:results}

\begin{comment}

\begin{figure}[ht]
\begin{center}
\includegraphics[width=1\linewidth]{img/AskUbuntu.png}
\caption{
MLP results on the AskUbuntu dataset vs. the dimensionality of HD vectors.
} 
\label{fig:AskUbuntu} 
\end{center}
\end{figure}

\begin{figure}[ht]
\begin{center}
\includegraphics[width=1\linewidth]{img/Chatbot.png}
\caption{
MLP results on the Chatbot dataset vs. the dimensionality of HD vectors.
} 
\label{fig:Chatbot} 
\end{center}
\end{figure}

\begin{figure}[ht]
\begin{center}
\includegraphics[width=1\linewidth]{img/WebApplication.png}
\caption{
MLP results  on the WebApplication dataset vs. the dimensionality of HD vectors.
} 
\label{fig:WebApplication} 
\end{center}
\end{figure}

\begin{figure*}[ht]
\begin{center}
\includegraphics[width=1.0\linewidth]{img/20NewsGroups.png}
\caption{
MLP results on the 20NewsGroups dataset vs. the dimensionality of HD vectors.
} 
\label{fig:20NewsGroups} 
\end{center}
\end{figure*}

\end{comment}

First, we report the results of the MLP classifier on all datasets as it represents a widely used class of algorithms -- neural networks.
The goal of the experiments was to observe how the dimensionality of HD vectors embedding $n$-gram statistics affects the $F_1$ scores and the computational resources. 
Figures~\ref{fig:AskUbuntu}-\ref{fig:20NewsGroups} present the results for the AskUbuntu, Chatbot, WebApplication, and 20NewsGroups datasets, respectively.  
The dimensionality of HD vectors varied as $2^k$, $k \in [5, 14]$.
All figures have an identical structure. 
Shaded areas depict $95$\% confidence intervals.
Left panels depict the $F_1$ score while right panels depict the train and test speed-ups as well as memory reduction. 
Note that there are different scales ($y$-axes) in the right panels.
A solid horizontal line indicates $1$ for the corresponding $y$-axis, i.e., the moment when both models consume the same resources.

The results in all figures are consistent in a way that up to a certain point $F_1$ score was increasing with the increasing dimensionality. 
For the small datasets even small dimensionalities of HD vectors (e.g., $32=2^5$) led to the $F_1$ scores, which are far beyond random. 
For example, for the AskUbuntu dataset, it was $84$\,\% of the conventional $n$-gram statistics $F_1$ score.
For the values above $512$ the performance saturation begins.  
Moreover, the improvements beyond $2048$ are marginal. 
The situation is more complicated for the 20NewsGroups dataset where for $32$-dimensional HD vectors $F_1$ score is fairly low though still better than a random guess ($0.05$).
However, it increases steeply until $1024$ and achieves its maximum at $4096$ being $92$\,\% of the conventional $n$-gram statistics $F_1$ score.
The dimensionalities above $4096$ showed worse results.

When it comes to computational resources, there is a similar pattern for all the datasets. 
The train/test speed-ups and memory reduction are diminishing with the increased dimensionality of HD vectors. 
At the point when the dimensionality of HD vectors equals the size of the conventional $n$-gram statistics, both approaches consume approximately the same resources.
These points in the figures are different because the datasets have different size of $n$-gram statistics: $3729$, $2753$, $2734$, and $192652$, for the AskUbuntu, Chatbot, WebApplication, and 20NewsGroups datasets, respectively.
Also, for all datasets, the memory reduction is higher than the speed-ups. 
The most impressive speed-ups and reductions were observed for the 20NewsGroups dataset (e.g., $186$ times less memory for $1024$-dimensional HD vectors). 
This is due to its large size it contains a huge number of $n$-grams resulting in large size of the $n$-gram statistics.
Nevertheless, even for small datasets, the gains were noticeable. 
For instance, for the WebApplication dataset at $256$ $F_1$ score was $99$\,\% of the conventional $n$-gram statistics while the train/test speed-ups and the memory reduction were $5.6$, $3.4$, and $7.6$, respectively.
Thus, these empirical results suggest that the quality of embedding w.r.t. the achievable $F_1$ score improves with increased dimensionality, however, after  saturation or peak increasing dimensionality further does not affect or worsen classification performance and becomes impractical when considering computational resources.

\begin{table}[t]
\scriptsize
\caption{ Performance of all classifiers for the AskUbuntu dataset.  }
\label{tab:classifiers:AskUbuntu}
\begin{center}
\begin{tabular}{|l|c|c|c||c|c|c||c|c|c|} 
%\hline
\cline{2-10}
 \multicolumn{1}{c}{} & \multicolumn{3}{|c||}{{$F_1$ score}} & \multicolumn{3}{|c||}{{Res.: SH vs. HD}} & \multicolumn{3}{|c|}{{Res.: SH vs. BPE}} \\ \hline %&
Class. & SH & BPE & HD & Tr. & Ts. & Mm. & Tr. & Ts. & Mm. \\ \hline %& Best Ind. \\ \hline
MLP                 & 0.92 & 0.91 & 0.91 & 4.62 & 3.84 & 6.18 & 1.67 & 1.61 & 1.72   \\ \hline%\hdashline %& 0.99 \\ \hdashline
PA      & 0.92 & 0.93 & 0.90 & 4.86 & 3.07 & 6.31 & 2.19 & 2.14 & 1.76   \\ \hline%\hdashline %& 0.97 \\ \hdashline
SGD       & 0.89 & 0.89  & 0.88 & 4.66 & 3.50 & 6.31 & 1.94 & 2.16 & 1.76  \\ \hline%\hdashline %& 0.99 \\ \hdashline
Ridge     & 0.90 & 0.91 & 0.90 & 3.91 & 4.74 & 6.31 & 1.63 & 1.62 & 1.76   \\ \hline%\hdashline %& 0.96 \\ \hdashline
KNN       & 0.79 & 0.72 & 0.82 & 2.11 & 4.53 & 8.48 & 1.56 & 1.79 & 1.76 \\ \hline%\hdashline %& 0.87 \\ \hdashline
NC    & 0.90 & 0.89 & 0.90 & 1.66 & 3.41 & 6.32 & 1.35 & 1.87 & 1.76 \\ \hline \hline %& 0.92 \\ \hdashline
LSVC          & 0.90 & 0.92 & 0.90 & 1.18 & 2.39 & 6.29 & 0.91 & 1.91 & 1.76  \\ \hline \hline %& 0.95 \\ \hdashline
RF      & 0.88 & 0.90 & 0.86 & 0.91 & 1.09 & 6.11 & 1.15 & 0.96 & 1.75  \\ \hline%\hdashline %& 0.95 \\ \hdashline
%Multinomial NB      & N/A & N/A &N/A     & N/A  & N/A \\ \hdashline %& 0.98 \\ \hdashline
BNB        & 0.91 & 0.92 & 0.85 & 2.30 & 3.72 & 6.34 & 1.96 & 2.42 & 1.76 \\ \hline %& 0.99 \\ 
\end{tabular} 
\end{center}
\end{table}

%Chatbot
\begin{table}[t]
\scriptsize
\caption{ Performance of all classifiers for the Chatbot dataset.  }
\label{tab:classifiers:Chatbot}
\begin{center}
\begin{tabular}{|l|c|c|c||c|c|c||c|c|c|} 
%\hline
\cline{2-10}
 \multicolumn{1}{c}{} & \multicolumn{3}{|c||}{{$F_1$ score}} & \multicolumn{3}{|c||}{{Res.: SH vs. HD}} & \multicolumn{3}{|c|}{{Res.: SH vs. BPE}} \\ \hline %&
Class. & SH & BPE & HD & Tr. & Ts. & Mm. & Tr. & Ts. & Mm. \\ \hline %& Best Ind. \\ \hline
MLP                 & 0.96 & 0.94 & 0.96 & 3.42 & 2.62 & 4.58 & 1.86 & 1.52 & 1.86     \\ \hline%\hdashline %& 0.99 \\ \hdashline
PA      & 0.95 & 0.91 & 0.94 & 4.40 & 2.38 & 4.72 & 2.29 & 2.22 & 1.92    \\ \hline%\hdashline %& 0.97 \\ \hdashline
SGD       & 0.93 & 0.93 & 0.92 & 3.16 & 2.06 & 4.72 & 1.88 & 1.84 & 1.92    \\ \hline%\hdashline %& 0.99 \\ \hdashline
Ridge     & 0.94 & 0.94 & 0.92 & 2.88 & 2.22 & 4.72 & 1.67 & 1.38 & 1.92     \\ \hline%\hdashline  %& 0.96 \\ \hdashline
KNN       & 0.75 & 0.71 & 0.83 & 1.66 & 3.59 & 6.51 & 1.43 & 1.79 & 1.92   \\ \hline%\hdashline %& 0.87 \\ \hdashline
NC    & 0.89 & 0.94 & 0.84 & 1.41 & 2.13 & 4.73 & 1.17 & 1.61 & 1.92   \\ \hline \hline %& 0.92 \\ \hdashline
LSVC          & 0.94 & 0.93 & 0.94 & 0.52 & 1.57 & 4.72 & 1.28 & 1.66 & 1.92  \\ \hline \hline  %& 0.95 \\ \hdashline
RF      & 0.95 & 0.95 & 0.91 & 0.95 & 1.10 & 4.61 & 1.16 & 0.98 & 1.91   \\ \hline%\hdashline  %& 0.95 \\ \hdashline
%Multinomial NB      & N/A & N/A &N/A     & N/A  & N/A \\ \hdashline %& 0.98 \\ \hdashline
BNB        & 0.93 & 0.93 & 0.82 & 1.92 & 2.60 & 4.73 & 1.53 & 1.72 & 1.92   \\ \hline %& 0.99 \\ 
\end{tabular} 
\end{center}
\end{table}

%WebApplication
\begin{table}[t]
\scriptsize
\caption{ Performance of all classifiers for the WebApplication dataset.}
\label{tab:classifiers:WebApplication}
\begin{center}
\begin{tabular}{|l|c|c|c||c|c|c||c|c|c|} 
%\hline
\cline{2-10}
 \multicolumn{1}{c}{} & \multicolumn{3}{|c||}{{$F_1$ score}} & \multicolumn{3}{|c||}{{Res.: SH vs. HD}} & \multicolumn{3}{|c|}{{Res.: SH vs. BPE}} \\ \hline %&
Class. & SH & BPE & HD & Tr. & Ts. & Mm. & Tr. & Ts. & Mm. \\ \hline %& Best Ind. \\ \hline
MLP                 & 0.77 & 0.77 & 0.79 & 3.10 & 2.00 & 4.43 & 1.74 & 1.44 & 1.73    \\ \hline%\hdashline %& 0.99 \\ \hdashline
PA       & 0.82 & 0.80 & 0.80 & 3.73 & 1.45 & 4.33 & 1.86 & 1.32 & 1.75   \\ \hline%\hdashline %& 0.97 \\ \hdashline
SGD       & 0.75 & 0.74 & 0.73 & 3.01 & 1.87 & 4.33 & 1.62 & 1.32 & 1.75  \\ \hline%\hdashline %& 0.99 \\ \hdashline
Ridge     & 0.79 & 0.80 & 0.80 & 1.66 & 2.40 & 4.34 &  0.71 & 1.09 & 1.75   \\ \hline%\hdashline %& 0.96 \\ \hdashline
KNN       & 0.72 & 0.75 & 0.76 & 1.16 & 2.76 & 5.96 & 1.14 & 1.51 & 1.76   \\ \hline%\hdashline %& 0.87 \\ \hdashline
NC    & 0.74 & 0.73 & 0.77 & 1.42 & 1.79 & 4.34 & 1.13 & 1.21 & 1.75  \\ \hline \hline %& 0.92 \\ \hdashline
 LSVC          & 0.82 & 0.80 & 0.80 & 1.04 & 1.48 & 4.29 & 0.47 & 1.18 & 1.75 \\ \hline \hline %& 0.95 \\ \hdashline
RF     & 0.87 & 0.85 & 0.72 & 0.95 & 1.26 & 4.11 & 1.05 & 1.12 & 1.73   \\ \hline%\hdashline %& 0.95 \\ \hdashline
%Multinomial NB      & N/A & N/A &N/A     & N/A  & N/A \\ \hdashline %& 0.98 \\ \hdashline
BNB        & 0.74 & 0.75 & 0.64 & 1.51 & 2.08 & 4.38 & 1.19 & 1.49 & 1.75  \\ \hline %& 0.99 \\ 
\end{tabular} 
\end{center}
\end{table}

\begin{table}[t]
\caption{Performance of all classifiers for 20NewsGroups.}
\label{tab:classifiers:20NewsGroups}
\begin{center}
\begin{tabular}{|l|c|c||c|c|c|} 
%\hline
\cline{2-6}
 \multicolumn{1}{c}{} & \multicolumn{2}{|c||}{{$F_1$ score}} & \multicolumn{3}{|c|}{{Resources: SH vs. HD}} \\ \hline %&
Classifier & SH & HD & Tr. sp.-up & Ts. sp.-up & Mem. red. \\ \hline %& Best Ind. \\ \hline
MLP                 & 0.72 & 0.64 & 53.23 & 79.50 & 93.19    \\ \hline%\hdashline %& 0.99 \\ \hdashline
PA       & 0.74 & 0.69 & 103.64 & 202.95 & 93.42  \\ \hline%\hdashline %& 0.97 \\ \hdashline
SGD       & 0.70 & 0.66 & 105.43 & 186.31 & 93.42   \\ \hline%\hdashline %& 0.99 \\ \hdashline
Ridge     & 0.16 & 0.71 & 45.46 & 338.01 & 93.42    \\ \hline%\hdashline %& 0.96 \\ \hdashline
KNN       & 0.31 & 0.31 & 184.47 & 65.87 & 127.54   \\ \hline%\hdashline %& 0.87 \\ \hdashline
NC    & 0.08 & 0.15 & 212.75 & 254.74 & 93.42 \\ \hline \hline %& 0.92 \\ \hdashline
LSVC          & 0.75 & 0.69 & 5.11 & 176.62 & 93.42 \\ \hline \hline %& 0.95 \\ \hdashline
RF      & 0.58 & 0.26 & 4.27 & 21.43 & 93.41   \\ \hline%\hdashline %& 0.95 \\ \hdashline
%Multinomial NB      & N/A & N/A &N/A     & N/A  & N/A \\ \hdashline %& 0.98 \\ \hdashline
BNB        & 0.60 & 0.15 & 57.72 & 56.54 & 93.42  \\ \hline %& 0.99 \\ 
\end{tabular} 
\end{center}
\end{table}

\begin{table}[t]
\scriptsize
\caption{Performance for AskUbuntu dataset with TF-IDF.  }
\label{tab:classifiers:AskUbuntu:TF}
\begin{center}
\begin{tabular}{|l|c|c|c||c|c|c||c|c|c|} 
%\hline
\cline{2-10}
 \multicolumn{1}{c}{} & \multicolumn{3}{|c||}{{$F_1$ score}} & \multicolumn{3}{|c||}{{TF vs. HD}} & \multicolumn{3}{|c|}{{TF-IDF vs. HD}} \\ \hline %&
Class. & TF & IDF & HD & Tr. & Ts. & Mm. & Tr. & Ts. & Mm. \\ \hline %& Best Ind. \\ \hline
MLP                  & 0.91 & 0.90 & 0.90 & 1.97 & 1.41 & 3.45 & 2.31 & 1.76 & 3.40   \\ \hline%\hdashline %& 0.99 \\ \hdashline
PA       & 0.93 & 0.93 & 0.90 & 3.58 & 2.15 & 3.48 & 3.57 & 2.51 & 3.50   \\ \hline%\hdashline %& 0.97 \\ \hdashline
SGD       & 0.90 & 0.89  & 0.86 & 3.81 & 4.25 & 3.48 & 3.32 & 3.98 & 3.50  \\ \hline%\hdashline %& 0.99 \\ \hdashline
Ridge     & 0.92 & 0.92 & 0.91 & 2.35 & 4.09 & 3.48 & 2.70 & 4.86 & 3.50   \\ \hline%\hdashline %& 0.96 \\ \hdashline
KNN       & 0.68 & 0.68 & 0.81 & 2.37 & 2.78 & 4.67 & 2.63 & 2.88 & 4.70 \\ \hline%\hdashline %& 0.87 \\ \hdashline
NC    & 0.88 & 0.86 & 0.89 & 2.77 & 3.50 & 3.48 & 2.63 & 3.56 & 3.50 \\ \hline \hline %& 0.92 \\ \hdashline
LSVC          & 0.94 & 0.93 & 0.91 & 2.07 & 2.54 & 3.47 & 1.93 & 2.81 & 3.49  \\ \hline \hline %& 0.95 \\ \hdashline
RF      & 0.89 & 0.88 & 0.84 & 0.87 & 1.08 & 3.38 & 0.93 & 1.11 & 3.40  \\ \hline%\hdashline %& 0.95 \\ \hdashline
%Multinomial NB      & N/A & N/A &N/A     & N/A  & N/A \\ \hdashline %& 0.98 \\ \hdashline
BNB        & 0.92 & 0.92 & 0.84 & 2.44 & 2.88 & 3.71 & 2.81 & 2.80 & 3.72 \\ \hline %& 0.99 \\ 
\end{tabular} 
\end{center}
\end{table}

 Tables~\ref{tab:classifiers:AskUbuntu}-\ref{tab:classifiers:20NewsGroups}\footnote{The notations Tr., Ts., Mm. in the tables stand for the train speed-up, test speed-up, and the memory reduction for the given classifier, respectively. SH stands for SemHash.} report the results for all datasets when applying all the considered classifiers. 
For the sake of brevity, a fixed dimensionality of HD vectors is reported only: $512$ for small datasets in Tables~\ref{tab:classifiers:AskUbuntu}-\ref{tab:classifiers:WebApplication} and $2048$ for the 20NewsGroups dataset in Table~\ref{tab:classifiers:20NewsGroups}. 
These dimensionalities were chosen based on the results in Figures~\ref{fig:AskUbuntu}-\ref{fig:20NewsGroups} as the ones allowing to achieve a good approximation of $F_1$ score while providing substantial speed-up/reduction.
We also performed experiments when using the BPE instead of the SemHash before extracting $n$-gram statistics.\footnote{ 
 Note that  Table~\ref{tab:classifiers:20NewsGroups} does not report the results for the BPE. 
This is purely due to high computational costs required to obtain the BPE model and vocabulary for this dataset. 
 }
Throughout the tables, the BPE demonstrated $F_1$ scores comparable to the SemHash while showing the train/test speed-ups and memory reduction at about $2$ times.
This is because the usage of the BPE resulted in smaller sizes of the $n$-gram statistics, which were $2176$, $1467$, and $1508$ for the AskUbuntu, Chatbot, and WebApplication datasets.   

In the case of HD vectors, the picture is less coherent. 
For example, there is a group of classifiers (e.g., MLP, SGD, KNN) where $F_1$ scores are well approximated (or even improved) while achieving noticeable computational reductions. 
In the case of LSVC, $F_1$ scores are well-preserved and there is $4-6$ times memory reduction but test/train speed-ups are marginal (even slower for training the Chatbot).
This is because LSVC implementation benefits from sparse representations (conventional $n$-gram statistics) while HD vectors in this study are dense. 
Last, for BNB and RF $F_1$ scores were not approximated well (cf. $0.93$ vs. $0.82$ for BNB in the case of the Chatbot). 
This is likely because both classifiers are relying on local information contained in individual features, which is not the case in HD vectors where information is represented distributively. 
The slow train time of RF is likely because in the absence of well-separable features it constructs large trees.

Due to the difference in the implementation (the official implementation of FastText only uses a linear classifier), we were not able to have a proper comparison of computational resources with the FastText.\footnote{We could have implemented the algorithm ourselves but it can be unfair to compare the resources, if we do not use the best practices unknown to us.}
However, we obtained the following $F_1$ scores with auto hyperparameter search: $0.91$, $0.97$, $0.76$ for the AskUbuntu, Chatbot, and WebApplication datasets, respectively.  
These results indicate that for the considered datasets there is no drastic classification performance improvement (even worse for the WebApplication) when using the learned representations of $n$-grams.

Table~\ref{tab:classifiers:AskUbuntu:TF} reports the results for the AskUbuntu dataset 
when applying all the considered classifiers on the features extracted with bag of words (denoted as TF) and TF-IDF.\footnote{We omitted the results for the WebApplication and  Chatbot datasets  due to the space limitations but the observed regularities were the same.}
 In these experiments as input to the classifiers we either used the features extracted by these methods or HD vectors ($N=512$) embedding these features.
% }
With respect to the compromise in terms of resources the classifiers performed similarly to the previous experiment with the difference that a typical speed-up and memory reduction were about three times for HD vectors.
When it comes to $F_1$ scores the results are consistent with the original motivation for the SemHash method, which argued that subword representations help in getting better performance compared to word-based representations at least for small datasets due to limited training data.

% \resizebox{\textwidth}{!}{%
% \begin{tabular}
% \end{tabular}
% }
% \resizebox{\linewidth}{!}{
\begin{table}[t]
\caption{$F_1$ score comparison of various platforms on three smaller datasets with results in the paper. }
\begin{center}
\begin{tabular}{|c|c|c|c|c|} 
\hline
Platform & Chat & Ask & WebApp & Ave. \\ [1ex] 
\hline
Botfuel  &  0.98 & 0.90 & 0.80 & 0.89\\ 
\hline%\hdashline     
Luis & 0.98 & 0.90 & 0.81 & 0.90\\ 
\hline%\hdashline      
Dialogflow  & 0.93 & 0.85 & 0.80 & 0.86\\
\hline%\hdashline      
Watson &  0.97 & 0.92 & 0.83 & 0.91\\
\hline%\hdashline      
Rasa  &  0.98 & 0.86 & 0.74 & 0.86\\
\hline%\hdashline      
Snips &  0.96 & 0.83 & 0.78  & 0.86\\ 
\hline%\hdashline      
Recast & \textbf{0.99} & 0.86 & 0.75 & 0.87\\
\hline%\hdashline
TildeCNN & \textbf{0.99} & 0.92 & 0.81  & 0.91\\
\hline \hline     
FastText & 0.97 & 0.91 & 0.76 & 0.88\\
\hline%\hdashline
SemHash & 0.96 & 0.92 & \textbf{0.87} & \textbf{0.92}\\
\hline%\hdashline
BPE & 0.95 & \textbf{0.93} & 0.85  & 0.91\\
\hline%\hdashline
HD vectors & 0.97 & 0.92 & 0.82  & 0.90\\
\hline
\end{tabular} 
\label{tab:f1 score comparison}
\end{center}
\end{table}
% }

Finally, for the small datasets Table~\ref{tab:f1 score comparison} places the results reported here in the context of results obtained in~\cite{SemHash19}\footnote{
Some results for Table~\ref{tab:f1 score comparison} are taken from~\cite{SemHash19}.
}. One thing to note in Table~\ref{tab:f1 score comparison} is the differences in the $F_1$ scores of the SemHash approach from the ones reported in~\cite{SemHash19} for all three small datasets. 
There were some data augmentation techniques, which were used in the paper, most prominently a QWERTY-based word augmentation accounting for the spelling mistakes.
This technique was not used here, which resulted in a slight difference in the obtained $F_1$ scores. 

\section{Discussion and conclusions}
\label{sec:disc}

The first observation is that the results on the 20NewsGroups dataset are not the state-of-the-art, which is currently $0.92$ $F_1$ score achieved with the BERT model as reported in~\cite{20NewsBERT}. 
Nevertheless, it is important to keep in mind that the main goal of the experiments with the 20NewsGroups dataset has been to demonstrate that $n$-gram statistics embedded into HD vectors allows getting the tradeoff even for a large text corpus. 
We even observed that for large datasets the usage of HD vectors is likely to provide the best gains in terms of resource-efficiency. 
Moreover, the gains on the small datasets were also noteworthy (several times).
Thus, based on these observations we conclude that HyperEmbed would be a very useful feature in the standard ML libraries.
It is also worth noting that an additional improvement for the inference step could be achieved when using binarized classifiers (see, e.g.,~\cite{shridhar2020end}). 
A more general conclusion is that it is worth revisiting results in the area of random projection~\cite{RPRachkovskij} as they are likely to allow achieving performance/resources tradeoff in a range of NLP scenarios (see, e.g.,~\cite{RandEmbed18} for one such example).

It was stated in Section~\ref{sec:classifiers} the speed-ups reported above did not include the time for forming HD vectors. 
The main reason for that is that our Python-based implementation of the method was quite inefficient, especially the cyclic shifts implemented with numpy.roll.
At the same time, as it could be seen from the formulation of the embedding method in Section~\ref{sec:HD:embed} its complexity is linear and depends on $n$ as well as on the length of the sample text, thus, fast implementation is doable. 
We made the proof-of-concept in Matlab, which is much faster. 
For example, for AskUbuntu forming $512$-dimensional HD vectors of the train split (the same machine) took $7.5$\,\% of the MLP training time, which is a positive result. 

Despite the demonstrated tradeoffs between the $F_1$ score and the computational resources, it is extremely hard to have an objective function, which would tell us when the compromise is acceptable and when it is not. 
In our opinion, a general solution would be to define a utility function, which would be able to assign a certain cost to both a unit of performance (e.g., $0.01$ increase in $F_1$ score) and a unit of computation (e.g., $10$ \% decrease in the inference time). 
The use of the utility function would allow deciding whether an alternative solution, which is, e.g.,  faster but less accurate, is better or not than the existing one. 
However, the main challenge here would be to define such a utility function since it would have to be defined for each particular application. 
Moreover, defining such functions even for the considered classification problems is out of the scope of this study. 
Nevertheless, we believe that it is the way forward to get an objective comparison criterion.

\bibliographystyle{IEEEtran}
\bibliography{IEEEfull}

\end{document}